\documentclass[letterpaper, 10 pt, conference]{ieeeconf}  
\IEEEoverridecommandlockouts                             
\usepackage{definitions}
\usepackage{abbreviations}
\usepackage{tikz}
\usepackage{flushend}
\usepackage{microtype}
\usepackage{adjustbox}

\newcommand{\GP}{\mathcal{GP}}
\DeclareMathOperator*{\argmax}{arg\,max}
\newcommand{\scF}{{\mathscr{F}}}

\newcommand{\scz}{{\mathscr{z}}}
\newcommand{\uscz}{{\underline{\scz}}}
\newcommand{\bep}{{\bm{\epsilon}}}
\newcommand{\ubep}{{\underline{\bep}}}
\newcommand{\Tbb}{\mathbb{T}}
\newcommand{\uio}{\underline{\iota}}
\newcommand{\rr}{{\boldsymbol{r}}}
\newcommand{\urr}{\underline{\rr}}

\newcommand{\ude}{\underline{\delta}}
\newcommand{\vc}{{\text{vec}}}
\newcommand{\md}{{\ttt{mod}}}
\newcommand{\ula}{{\underline{\lambda}}}
\newcommand{\ttt}[1]{{\texttt{#1}}}

\newcommand{\uscd}{{\underline{\mathscr{d}}}}
\newcommand{\bla}{\bm{\Lambda}}

\title{\LARGE \bf
	Gaussian Process on the Product of Directional Manifolds
}

\author{Ziyu~Cao and Kailai~Li 
	\thanks{The authors are with the Department of Electrical Engineering, Link\"oping University, Sweden. E-Mails: \tt{\{ziyu.cao, kailai.li\}@liu.se}}
}

\begin{document}
	
	\maketitle
	\thispagestyle{empty}
	\pagestyle{empty}

	\begin{abstract}
		We present a principled study on defining Gaussian processes (GPs) with inputs on the product of directional manifolds. A circular kernel is first presented according to the von Mises distribution. Based thereon, the hypertoroidal von Mises (HvM) kernel is proposed to establish GPs on hypertori with consideration of correlated circular components. The proposed HvM kernel is demonstrated with multi-output GP regression for learning vector-valued functions on hypertori using the intrinsic coregionalization model. Analytic derivatives for hyperparameter optimization are provided for runtime-critical applications. For evaluation, we synthesize a ranging-based sensor network and employ the HvM-based GPs for data-driven recursive localization. Numerical results show that the HvM-based GP achieves superior tracking accuracy compared to parametric model and GPs of conventional kernel designs.
	\end{abstract}
	
	\section{Introduction}\label{sec:introduction}
	Directional variables appear ubiquitously in control-related scenarios including signal processing, object tracking, and computer vision, among others~\cite{Ito2016,Glover2014,Zhe2018,ICRA20_Li,ECC20_Li,Fusion22_Li}. Typical examples of directional manifolds include the unit circle $\Sbb^1\subset\R^2$ and hyperspheres $\Sbb^{d-1}\subset\R^d$, which are by nature nonlinear and periodic. Many real-world applications, such as ground/aerial locomotion, robotic manipulation, and angle-of-arrival-based tracking, involve stochastic systems composing multiple directional inputs. Gaussian processes (GPs) have emerged as a powerful statistical tool for inferential modeling of uncertain dynamical systems~\cite{Rasmussen2006}, where dependence between system outputs at different inputs is captured by the similarity across the input domain quantified by the kernel. This paper focuses on the study of a representative case, namely GP modeling of vector-valued functions defined on the hypertorus $\Tbb^3\coloneqq\Sbb^1\times\Sbb^1\times\Sbb^1\subset\R^6$ (`$\times$' denotes the Cartesian product).
	
	The key challenge in addressing the considered problem is to design a suitable kernel quantifying the similarity between hypertoroidal data points adaptively to the underlying manifold structure. The investigation can be decomposed into two sub-problems. First, each circular component is inherently periodic and nonlinear, which cannot be simply handled by conventional kernels defined based on Euclidean distances. Second, it is essential to properly interpret the correlations among the circular components for informative similarity quantification. The former challenge can be overcome by transforming common periodic kernels w.r.t. the unit circle $\Sbb^1$~\cite{Rasmussen2006}. The latter refers to constructing kernels on a manifold composing multiple domains through the Cartesian product. In this regard, most existing solutions employ a straightforward strategy of multiplying the kernels belonging to each domain component, which overlooks the correlational interpretation between them and may result in information loss for inferential modeling~\cite{duvenaud2014automatic}.
	
	The major contribution of the presented work is a principled approach on establishing Gaussian processes on the product of directional manifolds. We propose the hypertoroidal von Mises (HvM) kernel for handling input data on hypertori with consideration of correlated circular components in a manifold-adaptive manner. Closed-form derivatives are provided for efficient hyperparameter optimization. Further, we synthesize an evaluation scenario for recursive tracking in sensor networks, where an unknown range sensing model is learned using angle-of-arrival data. Compared with parametric and GP modeling using common kernel designs, the proposed HvM-based GP enables superior tracking accuracy and robustness. We opensource our code under \texttt{github.com/ASIG-X/hyperToroidalGP}.
	
	The remainder of the paper is organized as follows. We summarize related works in \secref{sec:state}. Preliminaries on GP modeling are  introduced in \secref{sec:preliminaries}, after which the HvM kernel is proposed in \secref{sec:hvm}. We elaborate the evaluation in \secref{sec:evaluation}, and the work is concluded in \secref{sec:conclusions}.
	
	\section{Related Works}\label{sec:state}
	
	\subsubsection*{GPs on Riemannian Manifolds} In~\cite{Lindgren2011Explicit}, GP kernels on Riemannian manifolds (RMs) have been derived implicitly by solving stochastic partial differential equations incorporating the Laplace--Beltrami operator. An improvement on scalability has been achieved in~\cite{Borovitskiy2020Matern} via spectral theory for Mat\'{e}rn kernels on RMs. The approach was further extended in~\cite{Hutchinson2021Vector} to model tangential vector fields defined on RMs through projective geometry. These implicit strategies are generally applicable for developing RM-adaptive kernels, however, suffer from high theoretical complexity and computational expense, such as the eigendecomposition of Laplace-Beltrami operators, leading to a considerable gap for their usage in engineering practices. Alternatively, manifold-adaptive kernels can be explicitly designed \wrt the underlying manifold structure. This can lead to conciser procedures of establishing GPs on RMs. However, simply replacing the distance metric with the geodesic to generalize Gaussian kernels on RMs for non-Euclidean inputs does not guarantee positive definiteness. In this regard, Laplacian kernels are eligible if and only if the distance metric is conditionally negative definite~\cite{feragen2015geodesic}. For GP modeling on domains with complex but known structures, basic kernels of each domain component are typically combined through summation or multiplication~\cite{duvenaud2014automatic}. In~\cite{Majumdar2014}, GPs have been established on cylindrical surface ($\Sbb^1\times\R$) for spatiotemporal modeling of meteorological fields, where the kernel is obtained by multiplying a sinusoidal and a Laplacian kernel of $\mathscr{l}^1$-norm. In~\cite{Lang2014}, various GP kernels have been investigated on manifolds of unit quaternions and unit dual quaternions based on hyperspherical geometry on $\Sbb^3$ for spatial orientation/pose inferences.
	
	\subsubsection*{Directional Statistics} Distributions from directional statistics have provided valuable references for quantifying similarities between directional data points~\cite{Mardia2000directional}. Popular choices include the wrapped normal and the von Mises distributions on the unit circle, and the von Mises--Fisher distribution on hyperspheres~\cite{Fisher1953,Sensors21_Li,MFI20_Li}. For modeling directional quantities of antipodal symmetry, e.g., unit quaternions on $\Sbb^3\subset\R^4$, the Bingham distribution is commonly used and has been widely exploited in reasoning uncertain spatial orientations~\cite{gilitschenski2020deep,ECC19_Li}. As for distributions on composite directional domains, one important aspect is to interpret correlation across manifold components~\cite{Li2022Dissertation,Fusion20_Li}. Former works were shown on the torus $\Sbb^1\times\Sbb^1$ for temporal modeling of correlated wind directions~\cite{Kurz2015Toroidal} and on the manifold of unit dual quaternions for modeling uncertain spatial poses~\cite{Fusion19_Bultmann,LCSS21_Li,MFI19_Li}.
	
	\section{Gaussian Process Regression and Inference}\label{sec:preliminaries}
	\subsection{Gaussian Process Modeling}\label{subsec:gp}
	A Gaussian process is a collection of random variables on a certain domain. Any finite set of these random variables follows a multivariate Gaussian distribution. We denote a GP with scalar output as $\rr(\ux)\sim\GP(m(\ux), \sck(\ux, \ux'))$, with $m(\ux)$ and $\sck(\ux,\ux')$ being the mean and covariance functions of $\rr(\ux)$, respectively~\cite{Rasmussen2006}. Suppose there is an arbitrary function that is observed under uncertainty following $\rz=r(\ux)+\bep$ with noise $\bep\sim\mN(0,\sigma_r^2)$. Given a set of $n$ input locations and corresponding observations corrupted by noise, $\{(\ux_{\bullet,i},z_i)\}_{i=1}^n$, a GP yields a posterior of the function value at test locations $\{\ux_{\circ,i}\}_{i=1}^m$ in the form of a multivariate Gaussian distribution
	\begin{equation*}
		\urr_\circ\vert\{(\ux_{\bullet,i},z_i)\}_{i=1}^n,\{\ux_{\circ,i}\}_{i=1}^m\sim\mN(\hat{\ur}_\circ,\fC_\circ)\,,
	\end{equation*}
	with the mean and covariance given by
	\begin{equation}\label{eq:gpr}
		\hat{\ur}_\circ = \fK_{\circ\bullet}\fK^{-1}\uscz\,\eqand\fC_\circ=\fK_{\circ\circ}-\fK_{\circ\bullet}\fK^{-1}\fK_{\bullet\circ}\,,
	\end{equation}
	respectively~\cite{Rasmussen2006}, where $\fK=\fK_{\bullet\bullet}+\sigma_r^2\fI_n$. Element in matrix $\fK_{\bullet\circ}\in\R^{n\times{m}}$ at entry $(i,j)$ is the covariance given by the kernel evaluated at training and test locations $(\ux_{\bullet,i},\ux_{\circ,j})$. Matrices $\fK_{\bullet\bullet}\in\R^{n\times{n}}$, $\fK_{\circ\bullet}\in\R^{m\times{n}}$ can be obtained analogously. Vector $\uscz=[z_1, \cdots,z_n]^\top$ collects all observations, and $\fI_n\in\R^{n\times{n}}$ is an identity matrix. Further, the vector of predicted observations at test locations follows the distribution $\mN(\hat{\ur}_\circ,\fC_\circ+\sigma_r^2\fI_m)$.
	
	\subsection{Multi-Output Gaussian Processes}\label{subsec:mgp}
	For modeling vector-valued functions expressed as $\urz=\ur(\ux)+\ubep\in\R^d$, where $\ubep\sim\mN(\uzero_d,\fR)$ is the noise term and $\fR=\mdiag(\sigma_{r,1}^2,\cdots, \sigma_{r,d}^2)$ its associated covariances, multi-output Gaussian processes can be utilized. It takes the form $\urr(\ux)\sim\GP(\um(\ux),\scK(\ux, \ux'))$, with $\um(\ux)\in\R^d$ being a vector-valued function collecting the mean of each dimension and $\scK(\ux, \ux')$ the matrix-valued covariance function. To obtain a valid $\scK(\ux, \ux')$, one popular method is the intrinsic coregionalization model. This leads to
	\begin{equation}\label{eq:icm}
		\scK(\ux, \ux') = \sck(\ux, \ux') \fB\,,\eqwith\fB\in\R^{d\times{d}}
	\end{equation}
	being a coregionalization matrix that is positive semidefinite~\cite{Alvarez2012}. Given uncertain observations $\{\uz_i\}_{i=1}^n$ collected at training locations $\{\ux_{\bullet,i}\}_{i=1}^n$, the posterior at a test location $\ux_\circ$ follows 
	\begin{equation*}
		\urr_\circ\vert\{(\ux_{\bullet,i},\uz_i)\}_{i=1}^n,\ux_\circ\sim\mN(\hat{\ur}_\circ,\fC_\circ)\,.
	\end{equation*}
	Here, the mean and covariance are calculated according to the general form in \eqref{eq:gpr}, with components newly defined as $\fK_{\circ\bullet}=\fB\otimes\uk_{\circ\bullet}$, $\fK_{\circ\circ}= \sck(\ux_\circ,\ux_\circ)\fB$ and $\fK=\fB \otimes \fK_{\bullet\bullet}+\fR\otimes\fI_n$. Note that $\fK_{\bullet\bullet}$ still denotes the kernel matrix at training locations as given in \eqref{eq:gpr}. Furthermore, vector $\uscz$ is obtained through the vectorization $\uscz=\vc([\,\uz_1,\cdots,\uz_n\,]^\top)\in\R^{nd}$, and vector $\uk_{\circ\bullet}=[\sck(\ux_\circ,\ux_{\bullet,1}),\cdots,\sck(\ux_\circ,\ux_{\bullet,n})]$. Similarly to the scalar case, the predictive distribution of the observation at $\ux_\circ$ is obtained as $\urz_\circ\sim \mN(\hat{\ur}_\circ, \fC_\circ + \fR)$.
	
	\section{Gaussian Processes on the Hypertorus}\label{sec:hvm}	
	\subsection{A Circular Kernel Based on the von Mises Distribution}\label{subsec:vm}
\begin{figure}[t]
		\vspace{4mm}
		\centering
		\begin{tabular}{cc}
			\adjustbox{trim={0.26\width} {0.35\height} {0.23\width} {0.3\height},clip}{\includegraphics[width=0.31\textwidth]{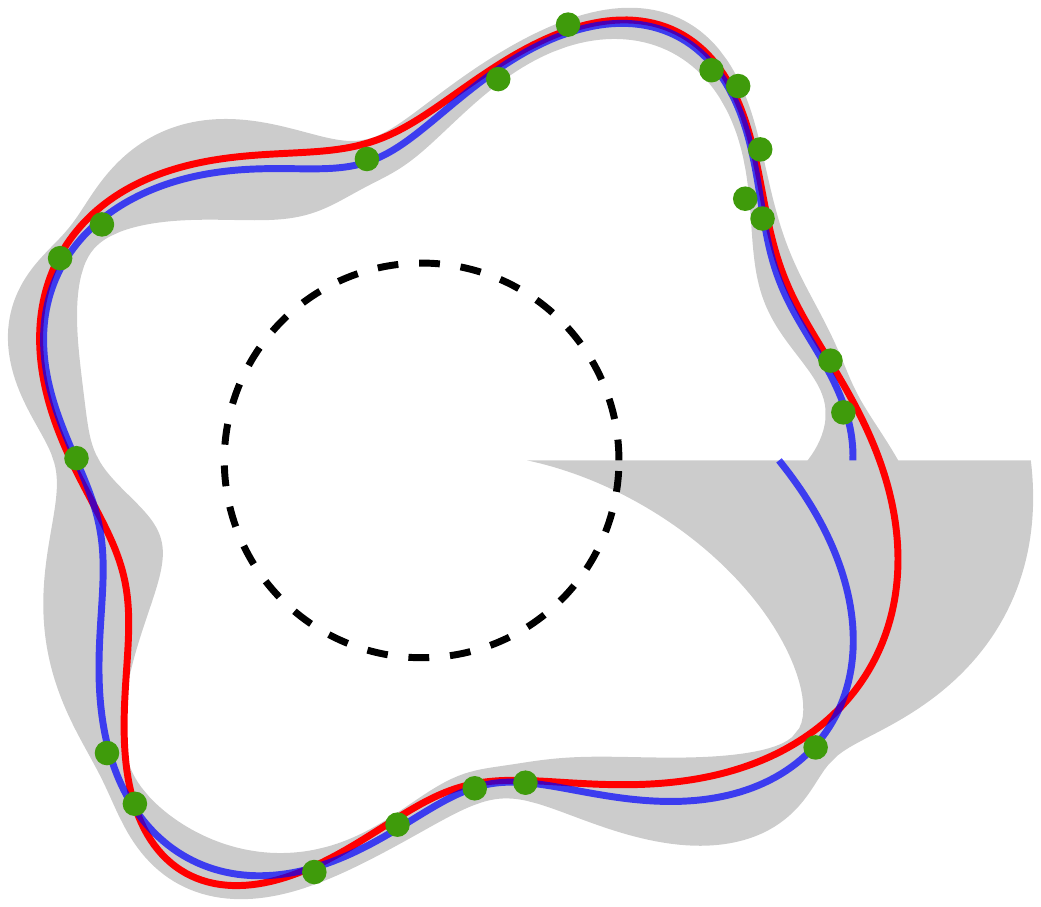}} &
			\adjustbox{trim={0.26\width} {0.35\height} {0.23\width} {0.3\height},clip}{\includegraphics[width=0.31\textwidth]{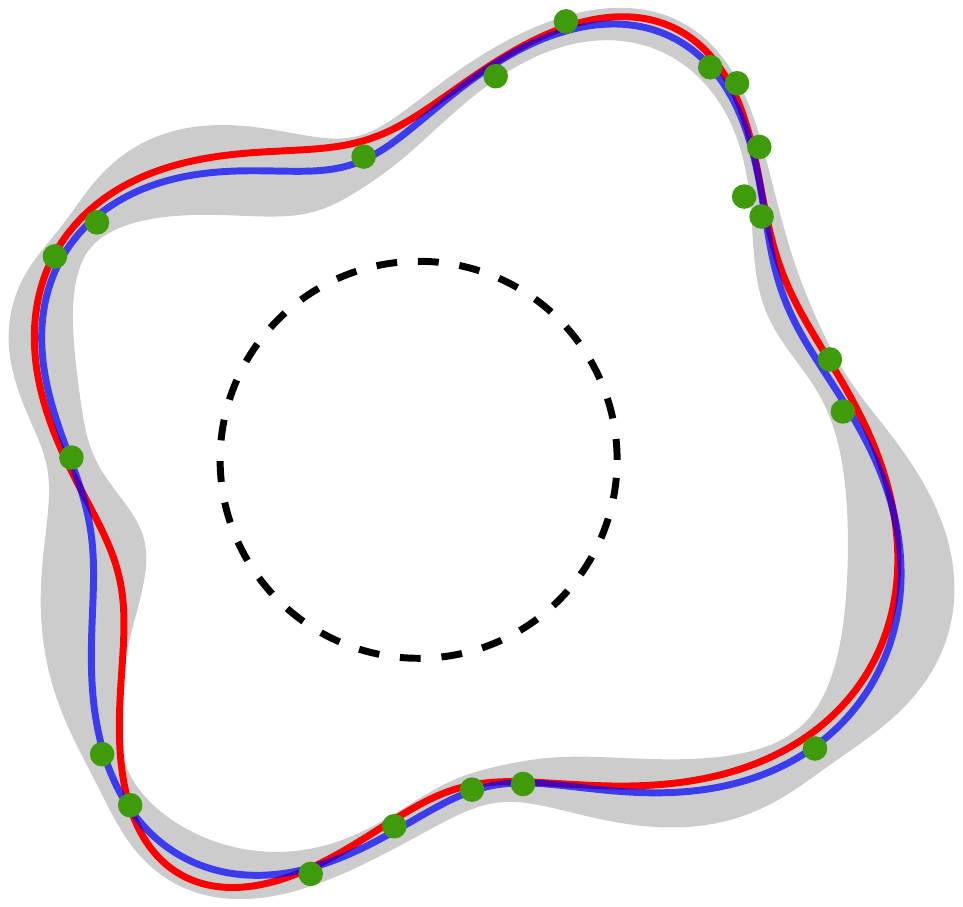}} \\
			(A) SE & (B) vM 
		\end{tabular}
		\caption{GP regression using different kernels on the circular function in \caseref{case1}. Given the same training set (green dots), the {vM} kernel produces a geometry-adaptive posterior (with blue curves and gray areas denoting means and variances, respectively), whereas the {SE} kernel fails.}
		\label{fig:s1gp}
		\vspace{-4mm}
	\end{figure}

	We now consider a scalar-valued function $r(\ux)$ with inputs defined on the unit circle, namely, $r:\Sbb^1\rightarrow\R$ for GP modeling. In order to quantify the similarity between two circular inputs $\uu,\uv\in\Sbb^1$, we formulate a kernel $\sck_\ttt{vM}(\uu,\uv)=\omega^2\exp(\lambda\,\uu^\top\!\uv)$ based on the von Mises distribution, with $\omega$ controlling signal variance and $\lambda> 0$ the concentration. Note that this kernel can be referred to as a reformulation of the periodic kernel with angular inputs~\cite[Eq.\,4.31]{Rasmussen2006}. To showcase its functionality within the GP framework compared with kernels on Euclidean domains, we demonstrate the following case study.
	\begin{Case Study}\label{case1}
		We synthesize a function on the unit circle by mixing three von Mises distributions and one Bingham distribution, with each component configured with individual parameters. The function is observed via $z=\frac{1}{3}\sum_{i=1}^3 f_\ttt{vM}^i(\ux)+f_\ttt{B}(\ux)+\epsilon$, with $\epsilon\sim\mN (0,0.0025)$ being additive noise. We embed the squared exponential ({SE}) kernel (distance metric $\mathscr{d}=\theta-\theta^\prime$, with $\theta$ and $\theta^\prime$ denoting the angular positions of $\uu$ and $\uv$, respectively) and the proposed von Mises (vM) kernel into the same GP regression scheme as introduced in \secref{subsec:gp}. Shown in \figref{fig:s1gp}-(A), the {SE} kernel produces discontinuous curves due to the aperiodic distance metric of Euclidean geometry. In contrast, the proposed {vM} kernel quantifies periodic similarity adaptively to circular geometry, inducing identical posteriors with period of $2\pi$ as plotted in \figref{fig:s1gp}-(B).
	\end{Case Study}
	
	\subsection{The Hypertoroidal von Mises Kernel}\label{subsec:hvm}
	We now aim to establish GPs on the hypertorus $\Tbb^3\subset\R^6$. For constructing kernels on the Cartesian product of multiple unit circles, a common strategy is to multiply kernels on each circular component~\cite{duvenaud2014automatic}. However, this only considers the similarity within each unit circle and neglects any potential correlations in between. To achieve informative GP modeling on hypertori, it is crucial to design a manifold-adaptive kernel with consideration of correlated circular components. For that, we define the hypertoroidal von Mises (HvM) kernel
	\begin{equation}\label{eq:hvm}
		\sck_\ttt{HvM}(\uu,\uv)=\omega^2\!\exp(\ula^\top\!\uscd(\uu,\uv)+\uscd(\uu,\uv)^\top\!\!\bla\uscd(\uu,\uv))\,,
	\end{equation}
	with $\uu,\uv\in\Sbb^1\times\Sbb^1\times\Sbb^1$ being a pair of hypertoroidal inputs. We express them component-wise w.r.t. each circle as 	$\uu\!=\![\,(\uu^1)^\top\!,(\uu^2)^\top\!,(\uu^3)^\top\,]^\top$ and $\uv\!=\![\,(\uv^1)^\top\!,(\uv^2)^\top\!,(\uv^3)^\top\,]^\top$. The metric $\uscd$ is defined as
	\begin{equation*}
		\uscd(\uu,\uv)=[\,(\uu^1)^\top\uv^1,(\uu^2)^\top\uv^2,(\uu^3)^\top\uv^3\,]^\top\,,
	\end{equation*}
	which measures distances between data points via inner products on each circular component. This is an analogous construction to the von Mises kernel given in \secref{subsec:vm}, and inherently guarantees the symmetry of function \eqref{eq:hvm} and the periodic nature of the hypertoroidal manifold.
	
	The proposed HvM kernel incorporates three hyperparameters, i.e., the $\omega\in\R$, $\ula\in\R^3$, and $\bla\in\R^{3\times3}$. Similar to the von Mises kernel in \secref{subsec:vm}, $\omega$ controls signal variance. The exponent of the kernel consists of a linear term and a quadratic term w.r.t. the distance metric $\uscd$. In the linear term, $\ula = [\,\lambda_1,\lambda_2,\lambda_3\,]^\top$ is a nonnegative vector and interprets concentrations on each circular component. In the second term, $\bla$ serves as a weighting matrix in the quadratic formulation and is defined as
	\begin{equation*}
		\bla=\bbmat0&a_1&a_3\\a_1&0&a_2\\a_3&a_2&0\ebmat\,,
	\end{equation*}
	with all elements being nonnegative. It is specifically designed to capture the correlations between the three pairs of circular components through the quadratic term. It allows for more informative similarity quantification of hypertoroidal data compared to simple products of circular kernels. To showcase this efficacy, we provide the following case study.
	
\begin{figure}[t!]
		\vspace{4mm}
		\centering
		\begin{tabular}{cc}  	 
			\adjustbox{trim={0\width} {0.22\height} {0.08\width} {0.255\height},clip}{\includegraphics[width=0.42\linewidth]{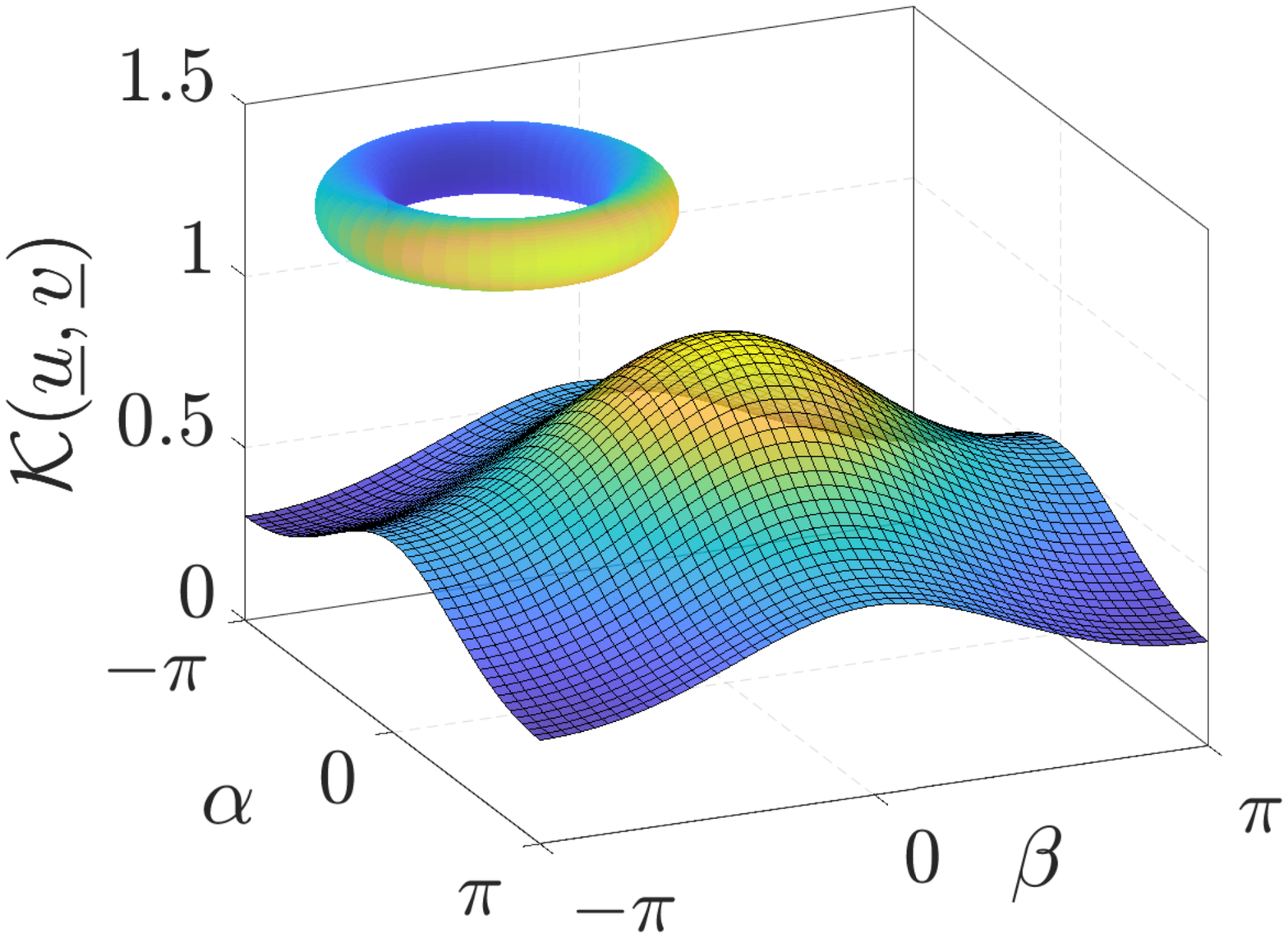}} &
			\adjustbox{trim={0\width} {0.22\height} {0.08\width} {0.255\height},clip}{\includegraphics[width=0.42\linewidth]{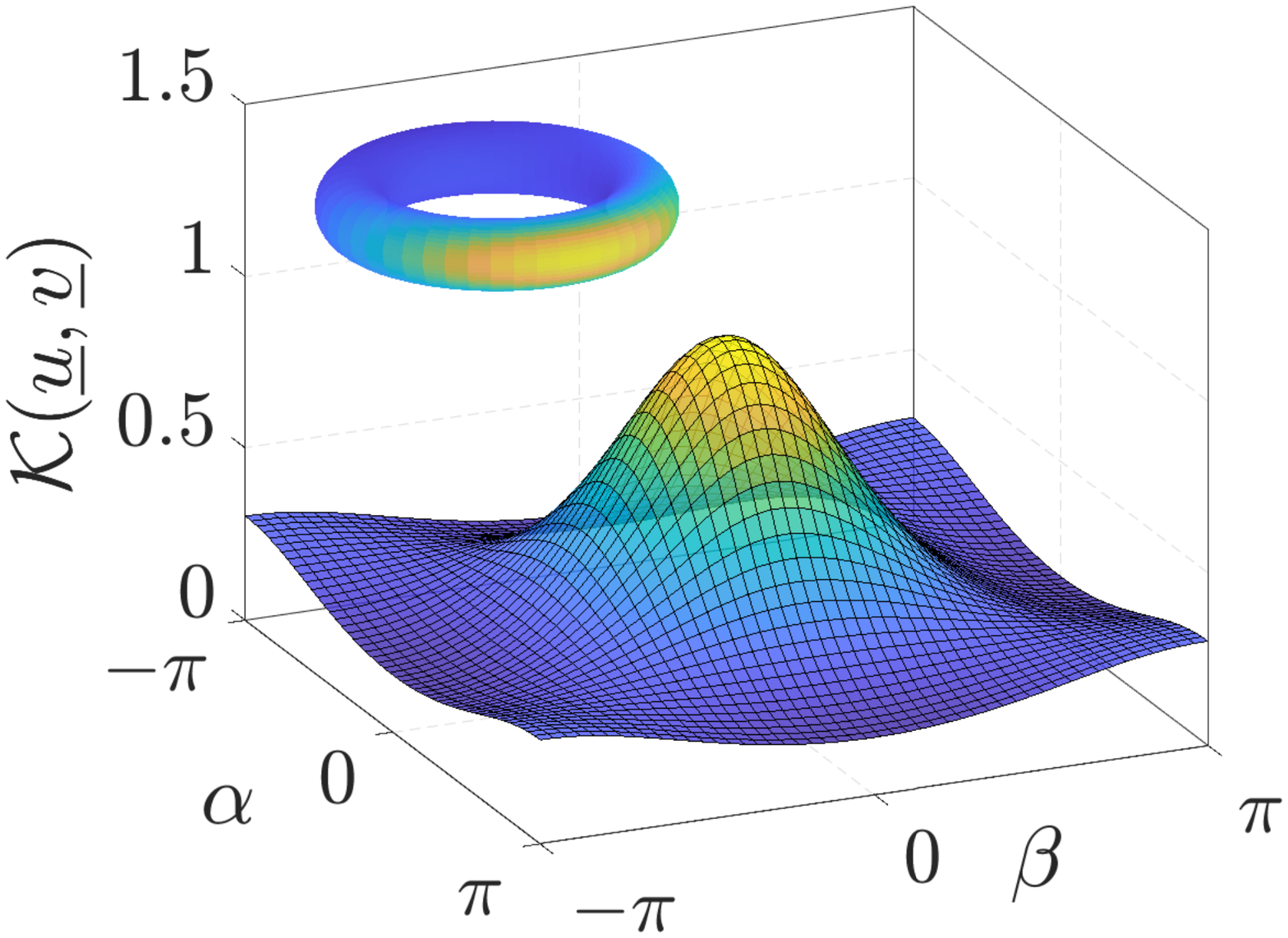}} \\
			(A)  & (B) \\
			\adjustbox{trim={0\width} {0.22\height} {0.08\width} {0.255\height},clip}{\includegraphics[width=0.42\linewidth]{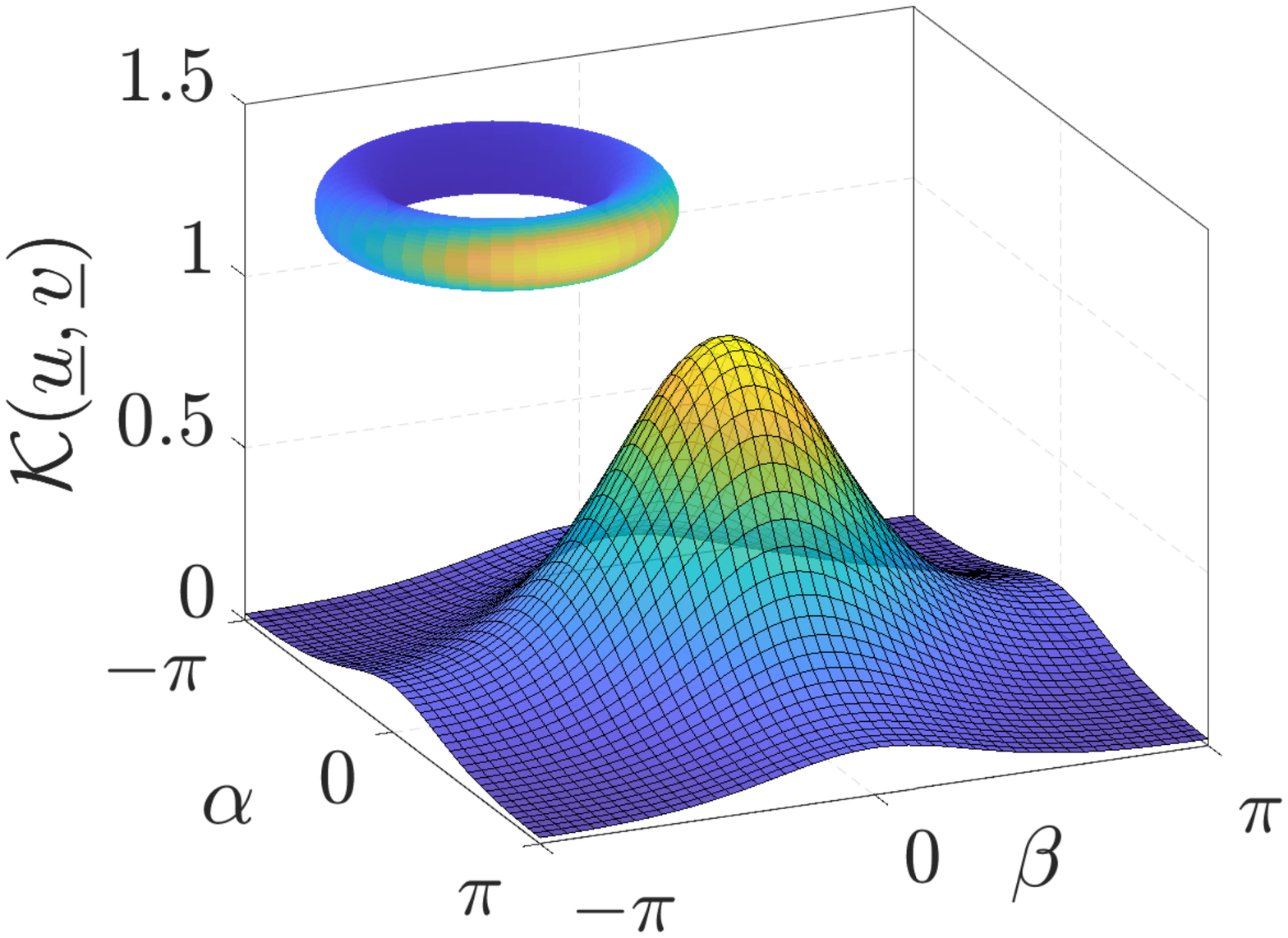}} & 		
			\adjustbox{trim={0\width} {0.22\height} {0.08\width} {0.255\height},clip}{\includegraphics[width=0.42\linewidth]{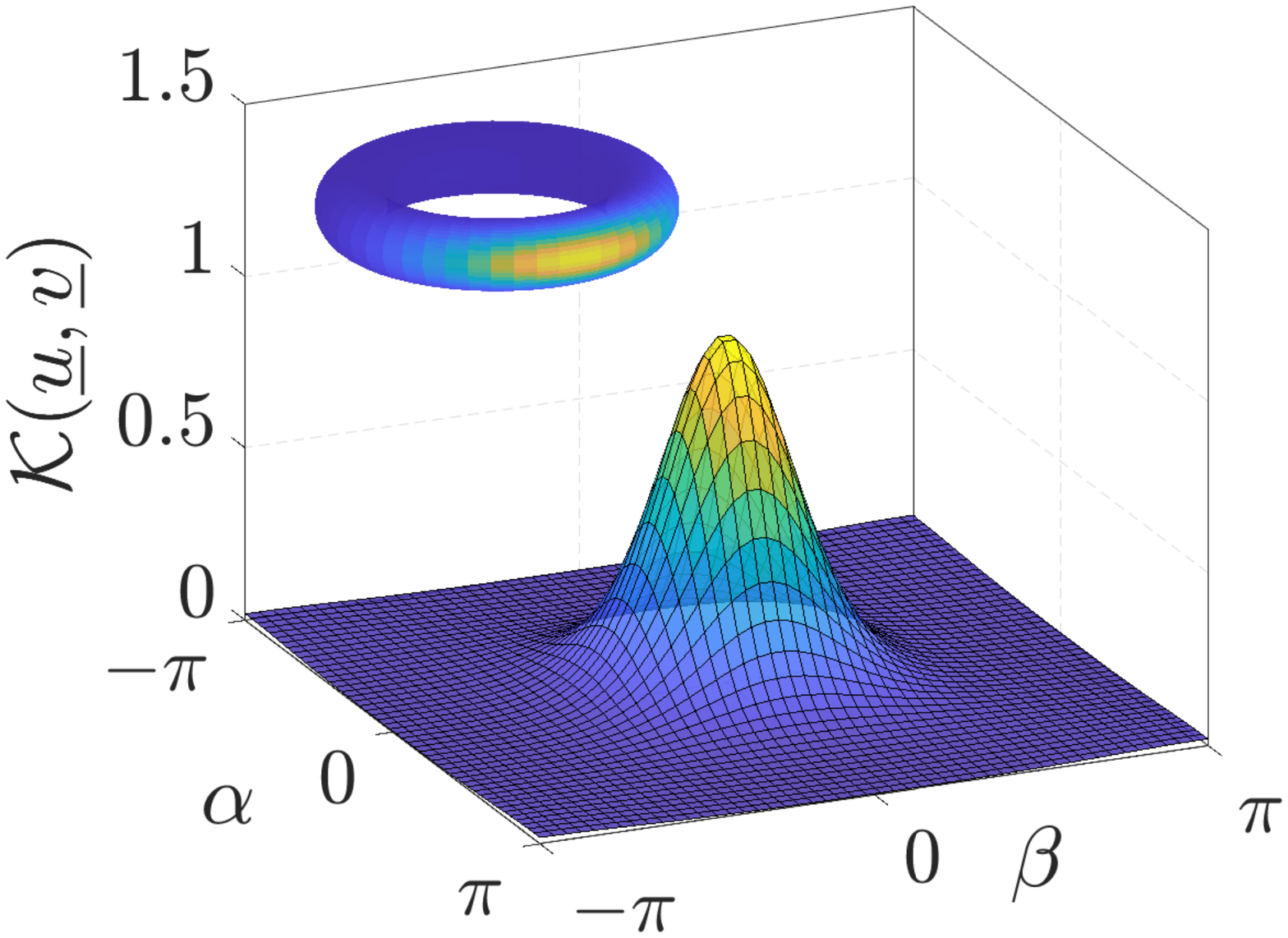}} \\			
			(C) & (D)
		\end{tabular}
		\caption{Proposed {HvM} kernels (normalized function values) configured in \caseref{case2} with mappings to tori.}
		\label{fig:hvm}
		\vspace{-2mm}
	\end{figure}

	\begin{Case Study}\label{case2}
		We configure the proposed kernel in \eqref{eq:hvm} \wrt the torus $\Sbb^1\times\Sbb^1\subset\R^4$ based on four sets of parameters $\{(\omega_i,\ula_i,\bla_i)\}_{i=1}^4$. These are as follows
		\begin{itemize}
			\item $\omega_1=\omega_2=\omega_3=\omega_4=1$\,,
			\item $\ula_1=\ula_2=[\,0.3,0.3\,]^\top$, $\ula_3= \ula_4=[\,1,1\,]^\top$\,, and 
			\item $\bla_1=\bla_{3} = \fzero_{2\times2}$, $\bla_2=\bbmat0&0.3\\0.3&0\ebmat$\,, $\bla_4=\bbmat0&1\\1&0\ebmat$\,.
		\end{itemize} 
		We evaluate the proposed kernel with one input fixed at $\uu=[\,1,0,1,0\,]^\top$, which corresponds to zero angular positions on each circle, and the other one given by
		\begin{equation*}
			\uv=[\,\cos(\alpha),\sin(\alpha),\cos(\beta),\sin(\beta)\,]^\top
		\end{equation*}
		w.r.t. angles $(\alpha,\beta)$. Shown in Fig. 2-(A) to (D), the resulted kernels are plotted corresponding to the parameter sets 1 to 4. In all cases, the kernel indicates the highest similarity at $(\alpha,\beta)=(0,0)$, namely, $\uu=\uv$\,. When $\bla = \fzero_{2\times2}$, as illustrated by \figref{fig:hvm}-(A) and (C), the proposed hypertoroidal kernel degenerates to a simple multiplication of von Mises kernels, which disregards the correlation between the two circular components. In contrast, the nonzero off-diagonal elements in $\bla_2$ and $\bla_4$ enable correlation interpretation between circles, leading to more distinguishable similarity quantification of function values as plotted in \figref{fig:hvm}-(B) and (D), respectively.
	\end{Case Study}
	
	\subsection{Hyperparameter Optimization}\label{subsec:opt}
	Varying the free parameters, e.g., those in the kernels, has a considerable impact on the performance of GP regression and inference. In general, these hyperparameters $\uth$ can be obtained via maximum likelihood estimation following $\uth^*=\argmax_{\uth}\scF(\uth)$. The objective is derived in the form of log marginal likelihood 
	\begin{equation}\label{eq:obj}
		\begin{aligned}
			\scF(\uth)&=2\log{p}\big(\uscz|\{\ux_{\bullet,i}\}_{i=1}^n,\uth\big)\\
			&=-\uscz^\top\fK^{-1}\uscz-\log\vert\fK\vert-n\log(2\pi)\,,
		\end{aligned}
	\end{equation}
	with $\fK$ and $\uscz$ specified according to \secref{sec:preliminaries}. In practice, the nonlinear optimization problem above is solved numerically with the kernel matrix evaluated in each iteration. The gradient of the objective \wrt the $i$-th element in $\uth$ takes the following general form 
	\begin{equation}\label{eq:grad}
		\pad{\scF(\uth)}{\theta_i}=\uscz^\top\fK^{-1}\pad{\fK}{\theta_i}\fK^{-1}\uscz -\operatorname{tr}\Big(\fK^{-1} \frac{\partial \fK}{\partial \theta_i}\Big)\,,
	\end{equation}	
	with $ \operatorname{tr}$ denoting the trace of a matrix.
	
	As for multi-output GPs using the proposed hypertoroidal von Mises kernel, the hyperparameters can be collected into $\uth=[\,\omega,\ula^\top\!,\ua^\top\!,\,\ub^\top\!,\usi_r^\top\,]^\top$. The first three components, $\omega$, $\ula$, and $\ua=[\,a_{1},a_{2},a_{3}\,]^\top\!$ are free parameters in \eqref{eq:hvm}. $\ub = \operatorname{vec}(\fB)$ denotes the vectorized coregionalization matrix in \eqref{eq:icm}. And $\usi_r=[\,\sigma_{r,1},\cdots,\sigma_{r,d}\,]^\top\!$ indicates observation noise variance on each output dimension according to \secref{subsec:mgp}.
	
	Computing \eqref{eq:grad} boils down to deriving the derivative of $\fK(\uth)$ \wrt hyperparameters in $\uth$, which we express element-wise now for the ease of exposition. The derivative of the kernel matrix \wrt the signal deviation $\omega$ follows
	\begin{equation*}
		\pad{\fK}{\omega}=\fB\otimes\pad{\fK_{\bullet\bullet}}{\omega}=\frac{2}{\omega}\,\fB\otimes\fK_{\bullet\bullet}\,,
	\end{equation*} 
	where $\fK_{\bullet\bullet}$ denotes the kernel matrix evaluated at training sets as introduced in \secref{subsec:mgp}. Further, the derivative of $\fK$ \wrt each element in the concentration vector $\ula$ follows
	\begin{equation*}
		\pad{\fK}{\lambda_s}=\fB\otimes\pad{\fK_{\bullet\bullet}}{\lambda_s}=\fB\otimes(\fK_{\bullet\bullet}\odot\fD^s_{\bullet\bullet})\,,
	\end{equation*}
	with $s\in\{1,2,3\}$ being the index of the $s$-th circular component of the hypertorus. Correspondingly, elements in $\fD^s_{\bullet\bullet}\in\R^{n\times{n}}$ follow $(\fD^s_{\bullet\bullet})_{ij}\coloneqq(\ux_{\bullet,i}^s)^\top\ux^s_{\bullet,j}$, which interprets distance between training inputs on each circle, and $\odot$ denotes the Hadamard product. As for differentiation \wrt the nonzero elements in matrix $\bla$, we have 
	\begin{equation*}
		\pad{\fK}{a_s}=\fB\otimes\pad{\fK_{\bullet\bullet}}{a_s}=2\fB\otimes\big(\fK_{\bullet\bullet}\odot\fD^s_{\bullet\bullet}\odot\fD^{s\,\md\,3+1}_{\bullet\bullet}\big)\,,
	\end{equation*}
	with $s\in\{1,2,3\}$. The derivative of $\fK$ \wrt the $s$-th element in the vectorized coregionalization matrix takes the form
	\begin{equation*}
		\pad{\fK}{b_s}=\pad{(\fB\otimes\fK_{\bullet\bullet})}{b_{s}}=\pad{\fB}{b_s}\otimes\fK_{\bullet\bullet}=\fE_{ij}\otimes\fK_{\bullet\bullet}\,,
	\end{equation*}	
	with $\fE_{ij}\in\R^{d\times{d}}$ denoting a matrix unit. Here, $b_s=(\fB)_{ij}$, with $s=d\,(j-1)+i\in\{1,\cdots,d^2\}$. The derivative of $\fK$ \wrt observation deviation $\sigma_{r,s}$ of each output domain $s\in\{1,\cdots,d\}$ can be derived in a similar fashion. It follows
	\begin{equation*}
		\pad{\fK}{\sigma_{r,s}}=\pad{(\fR\otimes\fI_n)}{\sigma_{r,s}}=2\sigma_{r,s}\fE_{ss} \otimes\fI_n\,,
	\end{equation*}
	where $\fE_{ss}\in\R^{d\times{d}}$ is a  matrix unit. In practice, we utilize the trust-regions method from \ttt{Manopt}~\cite{Manopt2014} for maximizing the objective in \eqref{eq:obj}.
	\begin{figure}[t]
		\vspace{4mm}
		\centering
		\begin{tabular}{ccc}
			\adjustbox{trim={0.23\width} {0.3\height} {0.27\width} {0.31\height},clip}{\includegraphics[width=0.57\linewidth]{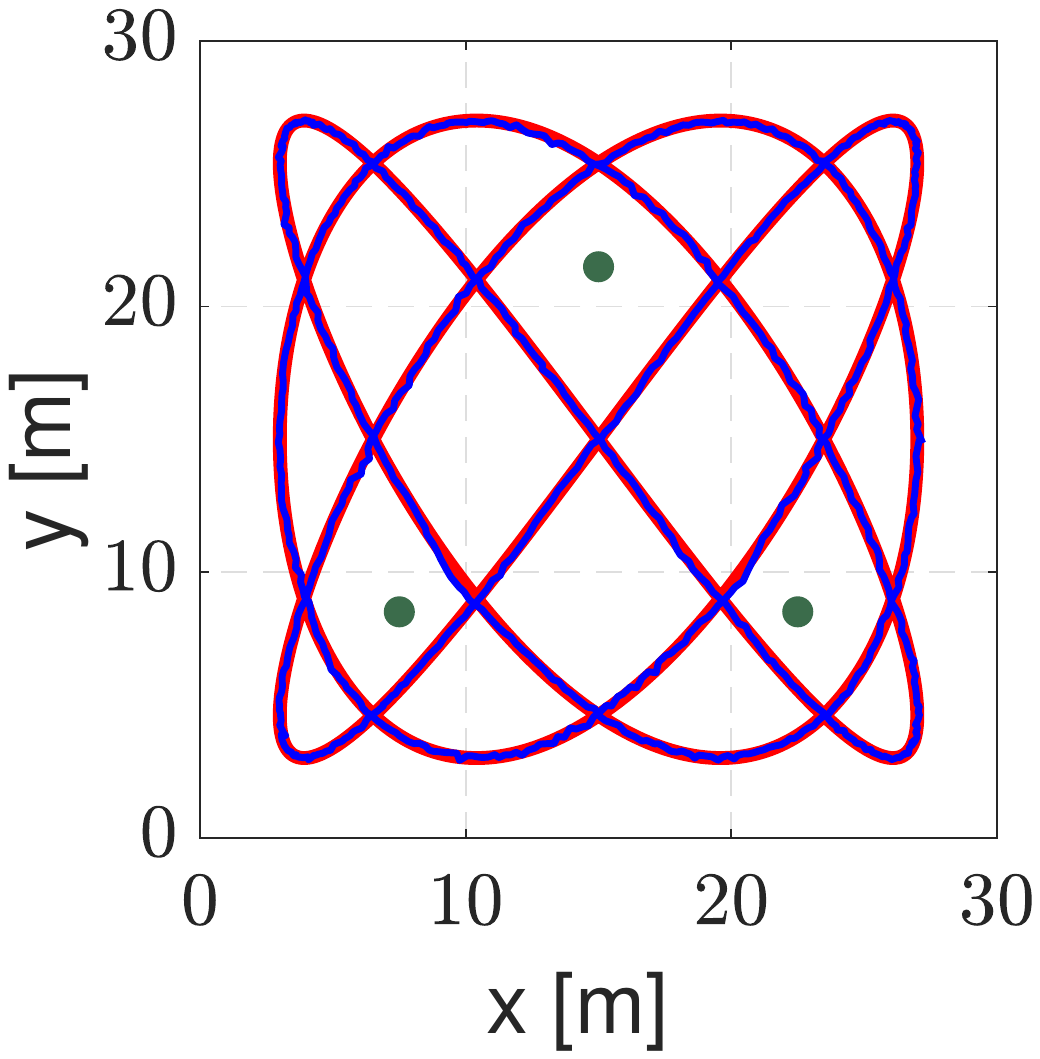}} &
			\adjustbox{trim={0.23\width} {0.3\height} {0.27\width} {0.31\height},clip}{\includegraphics[width=0.57\linewidth]{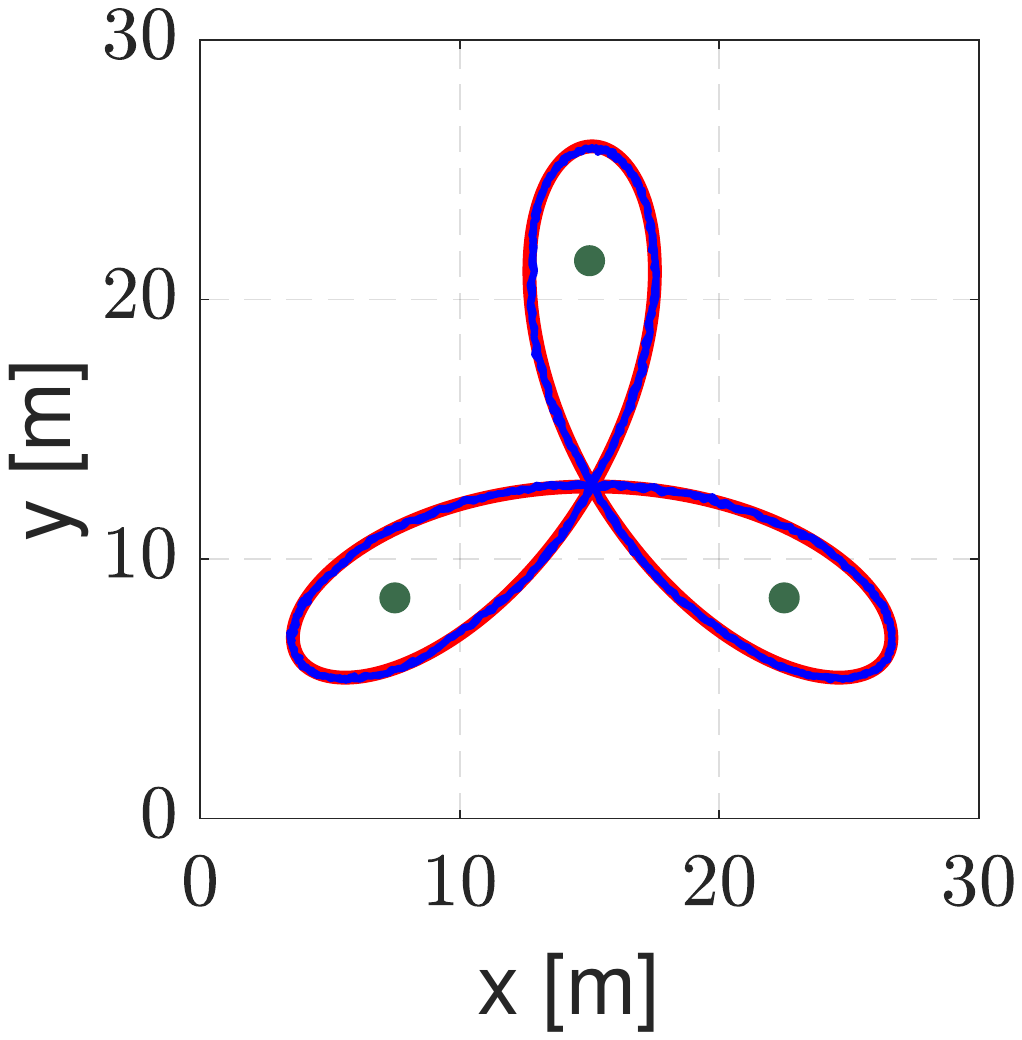}} &			
			\adjustbox{trim={0.23\width} {0.3\height} {0.27\width} {0.31\height},clip}{\includegraphics[width=0.57\linewidth]{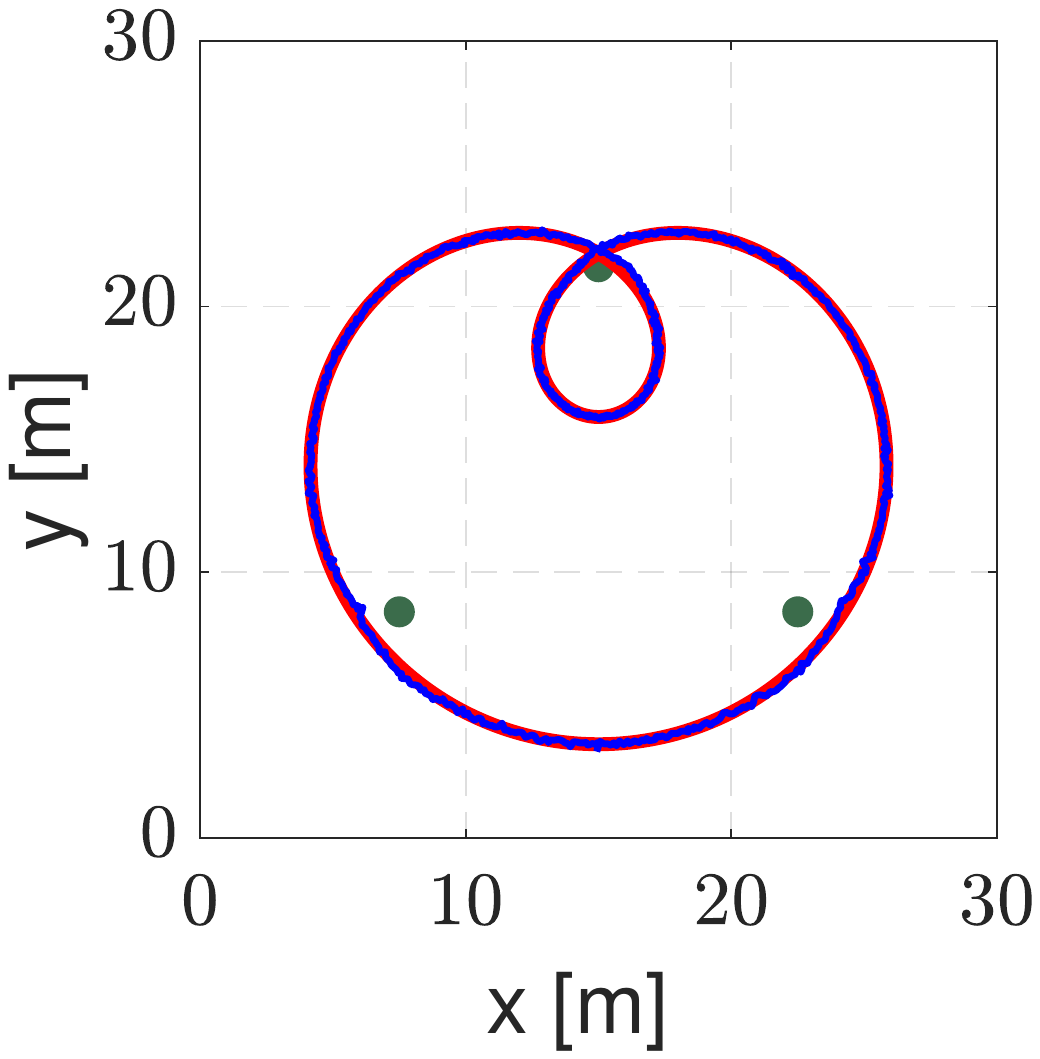}} \\	
			\ttt{T1} & \ttt{T2} & \ttt{T3}
		\end{tabular}
		\caption{Considered trajectories (red) and estimates (blue) given by GP-based particle filtering using the proposed HvM kernel. Noise level is set at $\xi=0.01$ (corresponding to sequences of \ttt{S1}). Green dots denote reference points.}
		\label{fig:exa}
		\vspace{-2mm}
	\end{figure}
	\begin{figure*}[t]
		\vspace{4mm}
		\centering
		\begin{tabular}{ccc}
			\adjustbox{trim={0.05\width} {0.0\height} {0.08\width} {0.05\height},clip}{\includegraphics[width=0.35\linewidth]{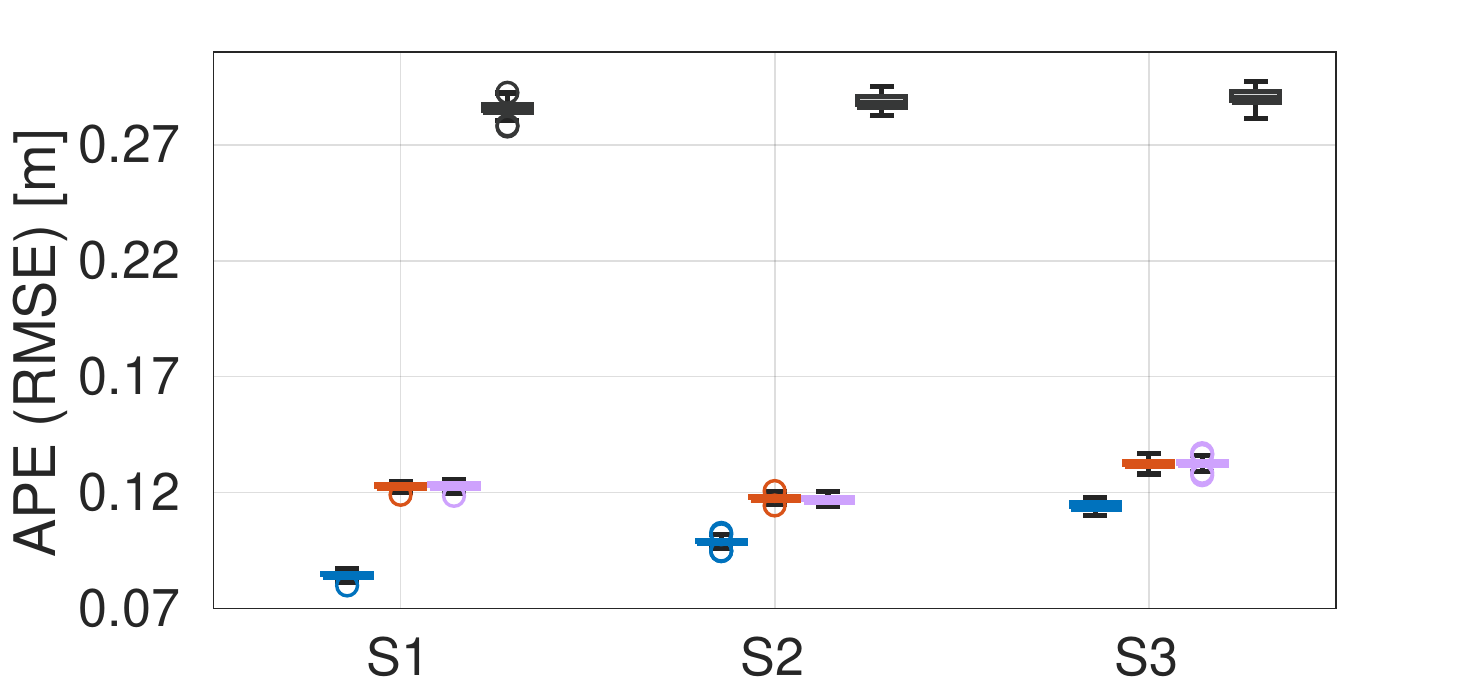}} &
			\adjustbox{trim={0.05\width} {0.0\height} {0.08\width} {0.05\height},clip}{\includegraphics[width=0.35\linewidth]{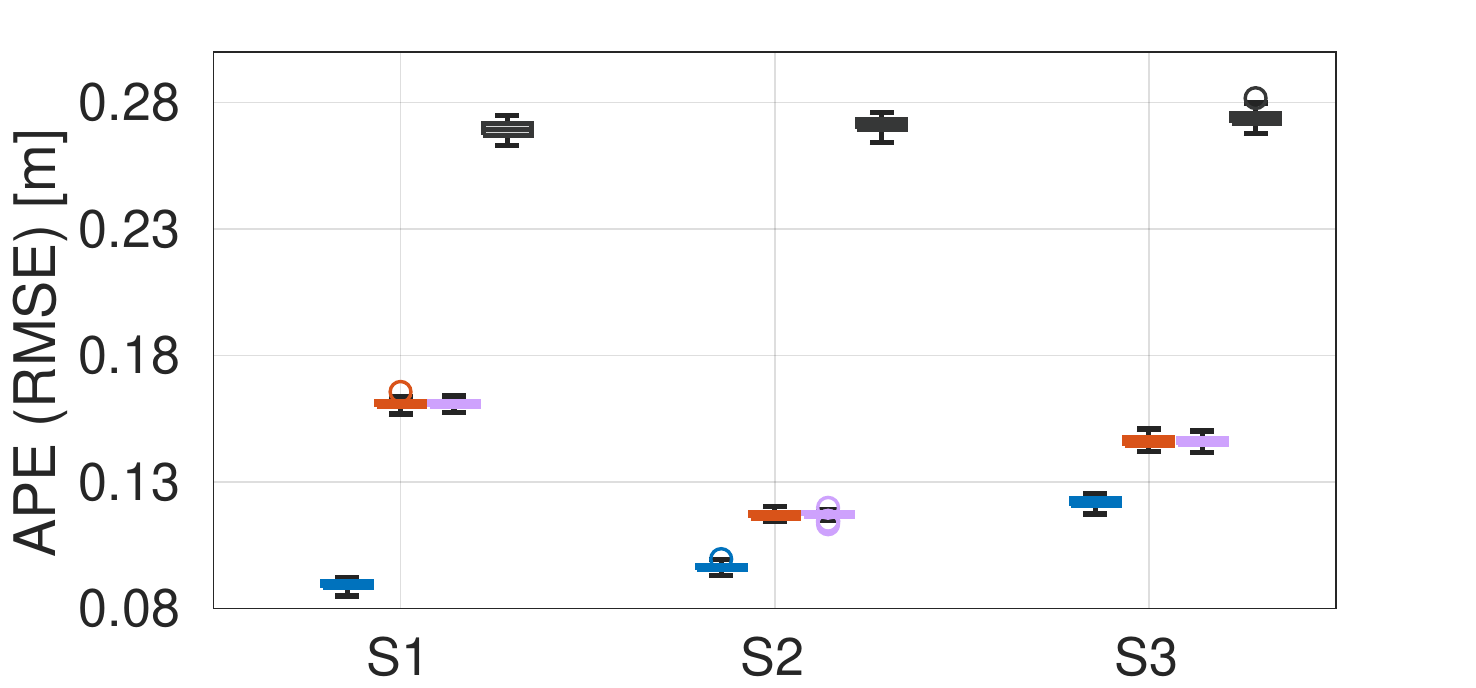}} &
			\adjustbox{trim={0.05\width} {0.0\height} {0.08\width} {0.05\height},clip}{\includegraphics[width=0.35\linewidth]{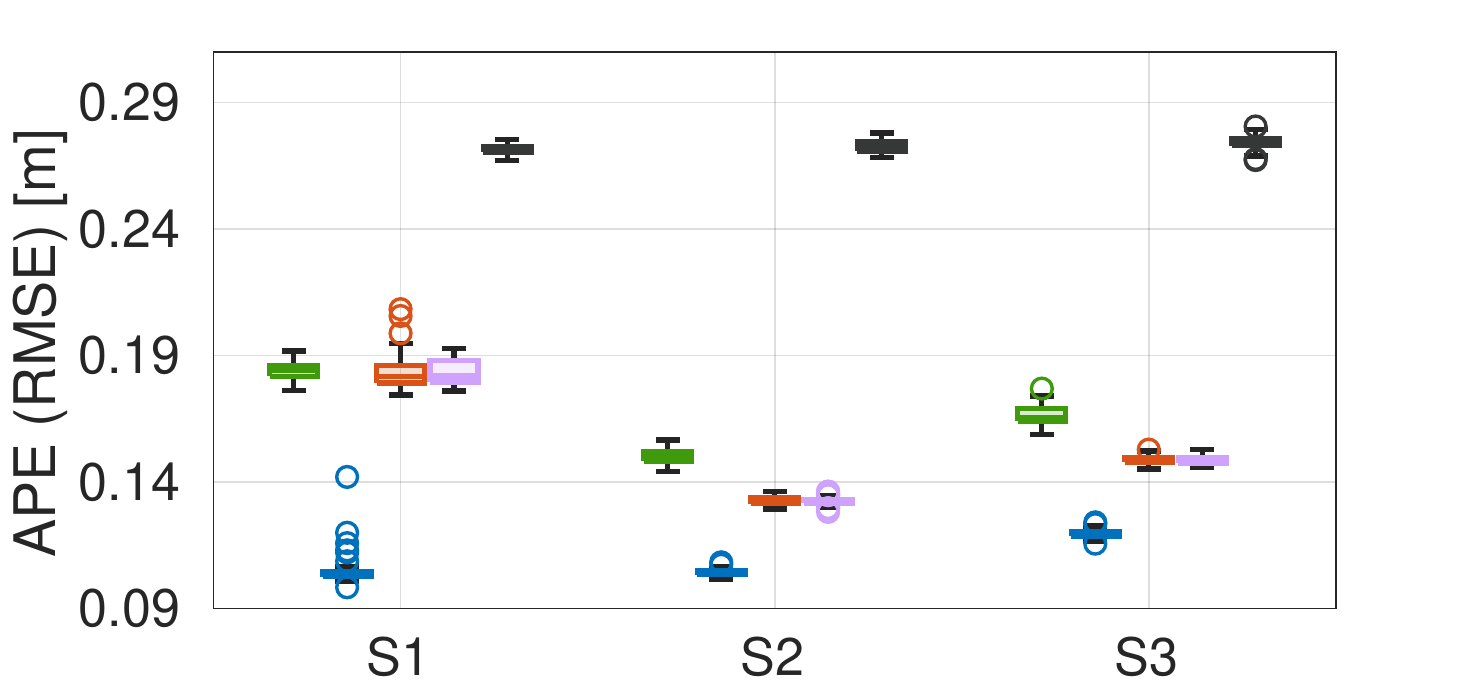}} \\
			\ttt{T1} & \ttt{T2} &\ttt{T3}
		\end{tabular}
		\caption{Results of particle filtering using parametric and different GP-based reweighting schemes in \secref{subsec:res}. RMSEs of the APEs are plotted for the trajectories $\ttt{T1}-\ttt{T3}$ using \ttt{boxchart} of default setting in \ttt{Matlab}. The proposed HvM-based GP (blue) reweighting consistently enables superior tracking accuracy compared to those based on {PvM} (red), and PPRD (lilac) kernels constructed via kernel product, whereas the parametric model (black) produces inferior performance with a considerable margin. The PSE kernel (green) leads to tracking failures in most sequences due to its nonperiodic definition.}
		\label{fig:error}
		\vspace{-4mm}
	\end{figure*}
	
	\section{Evaluation}\label{sec:evaluation}
	We now evaluate the proposed HvM-based GPs within particle filtering for data-driven recursive tracking in sensor networks. Here, an unknown range sensing model is learned on the hypertorus, which consists of multiple angle-of-arrival data as input.
	
	\subsection{Scenario Setup}\label{subsec:sce}
	We set up a two-dimensional area of $30\times30$ m$^2$, where a mobile agent is tracked while moving along different trajectories. We have the following system dynamics according to a random walk 
	\begin{equation}\label{eq:sys}
		\urx_{t+1}=\urx_{t}+\urw_{t}\,,
	\end{equation}
	with $\urx_t$, $\urx_{t+1}\in\R^2$ denoting positions and $\urw_t\in\R^2$ the process noise following $\urw_t\sim\mN(\uzero_2,\fQ_t)$. A range sensor, e.g., of ultrasonic or ultra-wideband modalities, is equipped onboard the platform observing distances to three reference points of coordinates $\{\uio_{s}^\ttt{r}\}_{s=1}^3\subset\R^2$ under uncertainty. This can be formulated as
	\begin{equation}\label{eq:meas}
		\urz_{t}=\uh(\urx_{t})+\ude_t+\urv_{t}\,,
	\end{equation}
	where $\urz_t\in\R^3$ and $\urv_t\in\R^3$ denote the range measurement and corresponding noise, respectively, and $\urv_t\sim\mN(\uzero_3,\fR_t)$. The observation function is given by 
	$\uh(\urx_t)\coloneqq[\Vert\uio_1^\ttt{r}-\urx_t\Vert,\Vert\uio_2^\ttt{r}-\urx_t\Vert,\Vert\uio_3^\ttt{r}-\urx_t\Vert]^\top$, which generates noise-free range readings of position state $\urx_t$ \wrt the three reference points. Moreover, a time-varying offset $\ude_t\in\R^3$ is added to the range signal to include possible interference in sensor networks, such as clock drift or signal reflection~\cite{manon2015}. For the sake of demonstration, we assume that it is proportional to the range, namely, $\ude_t=c\cdot\uh(\ux_t)$, with the ratio $c=0.05$.
	\begin{figure}[t]
		\centering
		\begin{tabular}{ccc}
			\adjustbox{trim={0.24\width} {0.3\height} {0.28\width} {0.32\height},clip}{\includegraphics[width=0.3\textwidth]{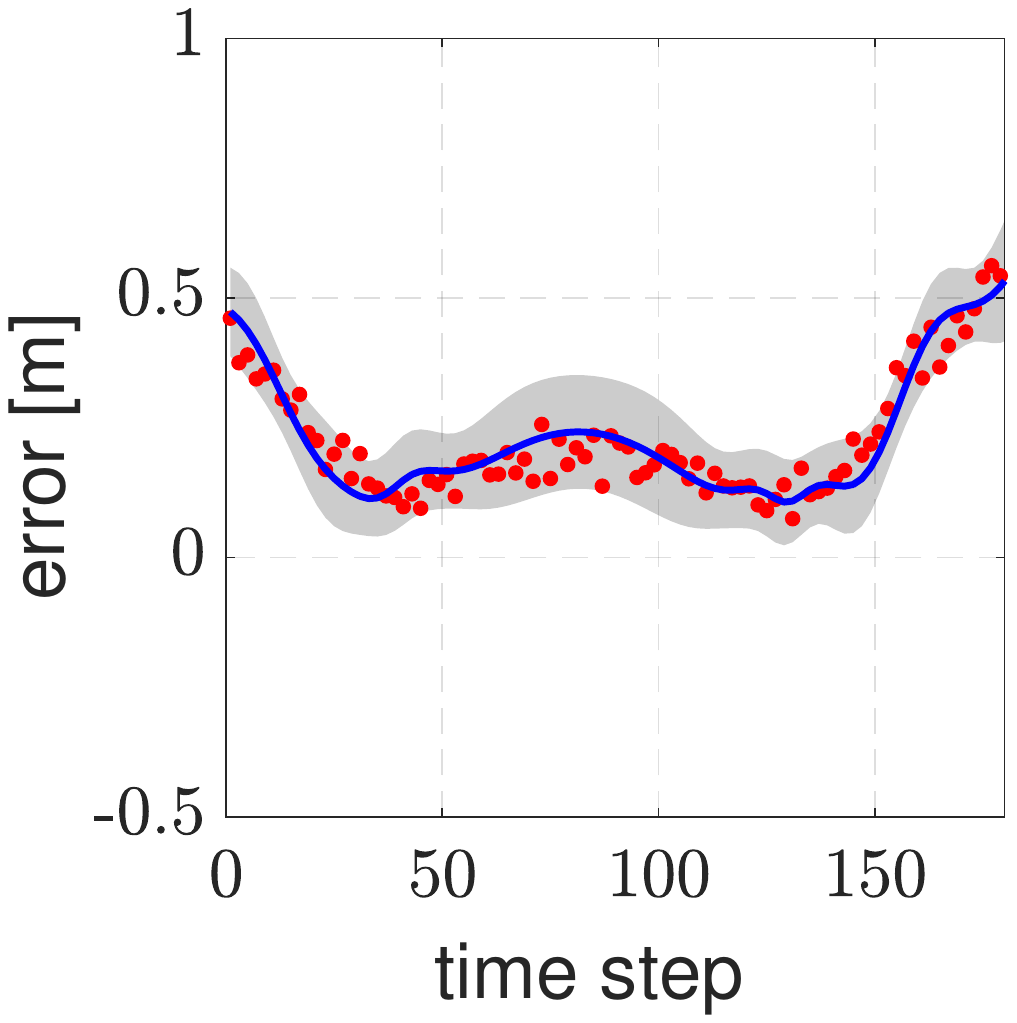}} &			
			\adjustbox{trim={0.28\width} {0.3\height} {0.28\width} {0.32\height},clip}{\includegraphics[width=0.3\textwidth]{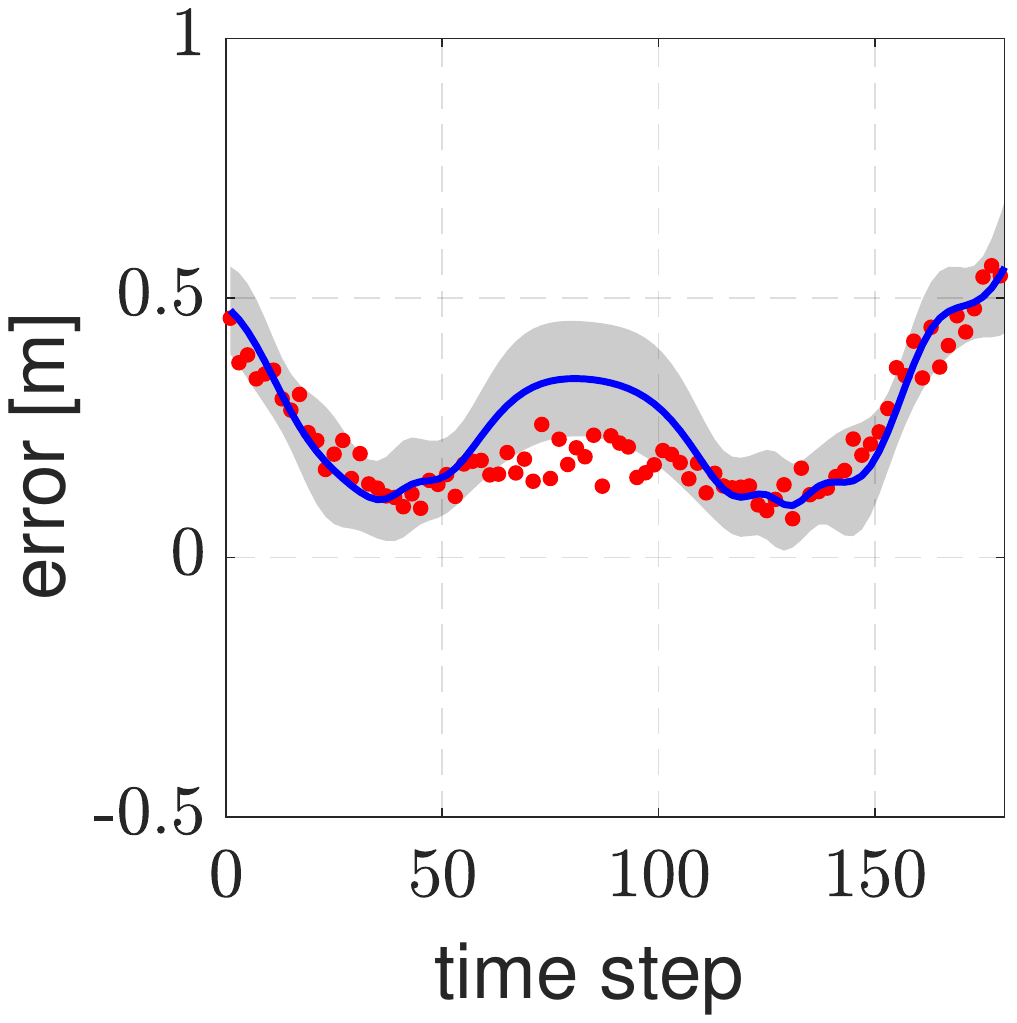}} &
			\adjustbox{trim={0.3\width} {0.3\height} {0.28\width} {0.32\height},clip}{\includegraphics[width=0.3\textwidth]{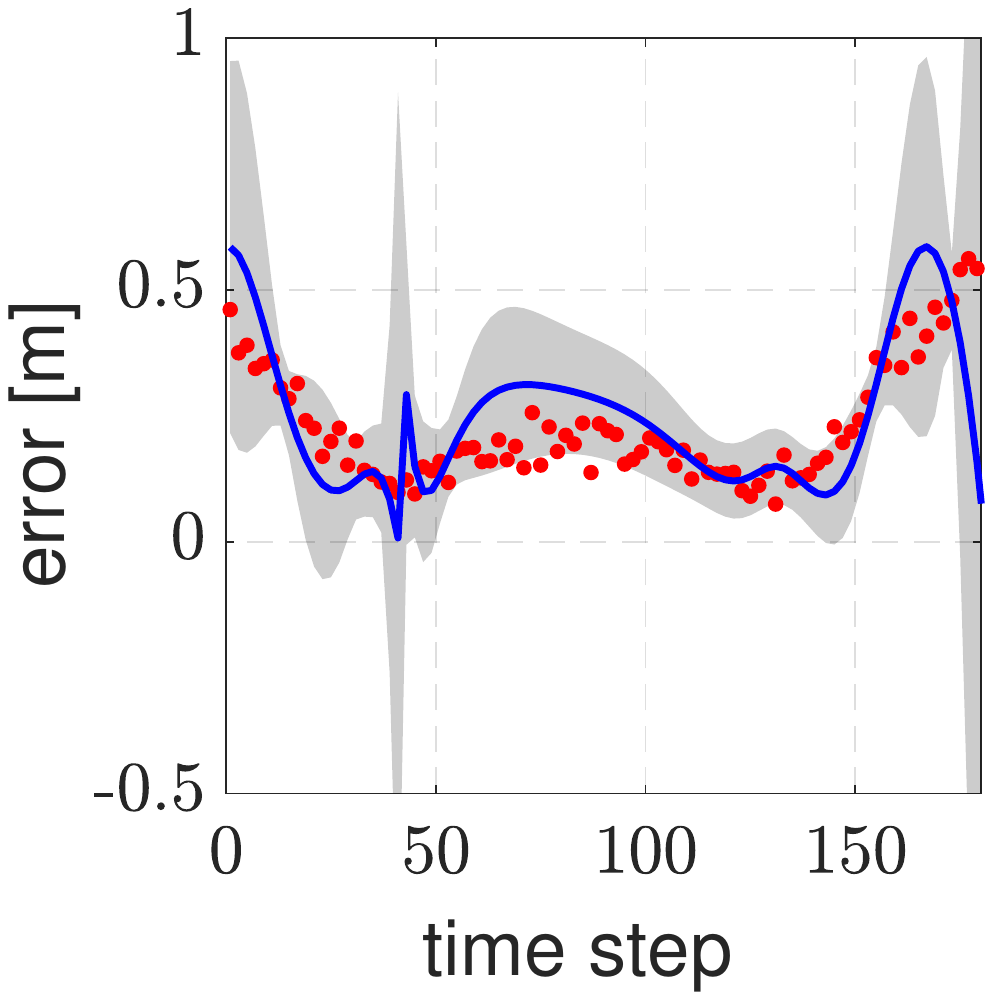}} \\
			(A) HvM  &(B) PvM &(C) PSE
		\end{tabular}
		\caption{Range measurements (red) and predictions (means and variances plotted by blue curves and gray areas, respectively) \wrt ground truth over \ttt{T2} of noise level \ttt{S2}. The reference point on the top in \figref{fig:exa} is selected for demonstration.}
		\label{fig:comp}
		\vspace{-4mm}
	\end{figure}
	
	\subsection{Multi-Output GPs on Hypertorus}\label{subsec:training}
	Suppose that the offset and uncertainty in range observations are unknown. However, we are able to obtain accurate angle-of-arrival (AoA) readings from each reference point \wrt $24 \times 10$ grid points $\{\ux_i\}_{i=1}^{n=240}$ that are uniformly spaced on the $xy$-plane. Meanwhile, range measurements given by the onboard sensor are collected, inducing a training set of $\{(\ual_i,\uz_i)\}_{i=1}^{n}$, where the range $\uz_i\in\R^3$. The input $\ual_i=[\,(\ual_i^1)^\top\!,(\ual_i^2)^\top\!,(\ual_i^3)^\top\!\,]^\top\!\in\Tbb^3\!\subset\R^6\!$ incorporates the AoA signal of each reference point \wrt grid point $\ux_i$, which can be obtained via
	\begin{equation}\label{eq:aoa}
		\ual_i^s=[\,(\uio_{s}^\ttt{r}-\ux_i)_\ttt{x},(\uio_{s}^\ttt{r}-\ux_i)_\ttt{y}]^\top/{\Vert\uio_{s}^\ttt{r}-\ux_i\Vert}\in\Sbb^1\,,
	\end{equation}
	with $s\in\{1,2,3\}$. Such a scenario occurs commonly when there exists no positioning system (e.g., camera networks) covering the whole tracking space, whereas certain portable devices such as total stations can be easily deployed to provide accurate relative angles between two locations~\cite{aoa1999}. Based on the training set, we exploit the proposed HvM-based vector-valued GPs to learn the unknown range sensing model as introduced in \secref{sec:hvm}. According to \secref{subsec:mgp}, we reformulate the measurement model in \eqref{eq:meas} into $\urz_t=\ur(\ual_t)+\urv_t$, with $\ual_t\in\Tbb^3$ incorporating the AoA inputs and $\urv_t$ the uncertainty in the output. Given the scenario setup in \secref{subsec:sce}, the noise-free range observation follows $\ur(\ual_t)=h(\urx_t)+\ude_t = (1+c)\uh(\ux_t)$.
	
	\subsection{GP-Based Particle Filtering}\label{subsec:filter}
	We apply particle filtering to track the mobile agent set up in \secref{subsec:sce}~\cite{fredrik2002}. Given a particle $\hux_{t\vert{t-1}}$ drawn at timestamp $t$ from the prior estimate (particle index is omitted for brevity), we first compute the corresponding hypertoroidal state $\hat{\ual}_{t\vert{t-1}}\in\Tbb^3$ following \eqref{eq:aoa}. Here, each circular component is given by $\hat{\ual}_{t\vert{t-1}}^s=[\,(\uio_{s}^\ttt{r}-\hux_{t\vert{t-1}})_\ttt{x},(\uio_{s}^\ttt{r}-\hux_{t\vert{t-1}})_\ttt{y}]^\top/{\Vert\uio_{s}^\ttt{r}-\hux_{t\vert{t-1}}\Vert}$ \wrt each reference point indexed by $s\in\{1,2,3\}$. We further inquire the pre-trained GPs (using training data $\{(\ual_{i},\uz_i)\}_{i=1}^{n}$) at $\hat{\ual}_{t\vert{t-1}}$ to predict corresponding range measurement distribution, i.e., $\urz_t\vert\{(\ual_{i},\uz_i)\}_{i=1}^{n},\hat{\ual}_{t\vert{t-1}}\sim\mN(\hat{\ur}_{t\vert{t-1}},\fC_{t\vert{t-1}} + \fR)$,
	where $\hat{\ur}_{t\vert{t-1}}$ and $(\fC_{t\vert{t-1}} + \fR)$ are posterior mean and covariance given by the GP regression in~\secref{subsec:mgp}, respectively. Based thereon, the likelihood function is directly obtained for reweighting the prior particle $\hux_{t\vert{t-1}}$ given current measurement $\uz_{t}$. Afterward, particles are updated via resampling.
	
	\subsection{Evaluation and Results}\label{subsec:res}
	We equip the GP-based particle filter in \secref{subsec:filter} with the proposed {HvM} kernel. For comparison, we instrument the product of squared exponential kernels (PSE), the product of periodic kernels (PPRD), and the product of von Mises kernels (PvM) on the hypertorus through multiplication of the corresponding circular kernels~\cite{duvenaud2014automatic}. GPs based on different kernels are trained using the same data set collected as introduced in \secref{subsec:training}. Furthermore, we approximate range measurement noise with a three-dimensional Gaussian distribution using the same training data to provide likelihood functions for a parametric form. We set the process noise in \eqref{eq:sys} with covariance $\fQ_t=\mdiag{(0.16,0.16)}$ and the measurement noise in \eqref{eq:meas} with covariance $\fR_t=\xi^2\fI_3$, where $\xi\in\{0.01,0.03,0.05\}$ controls the level of uncertainty. We deploy $100$ particles to all particle filters for $1000$ time steps and perform evaluation based on $100$ Monte Carlo runs for different noise levels and trajectories shown in \figref{fig:exa}. Subsequently, we compute the absolute position error (APE) to quantify the tracking accuracy.
	
	Shown in \figref{fig:error}, we collect the root-mean-square error (RMSE) of APEs given by the particle filters configured above on sequences of varying trajectories (\ttt{T1}-\ttt{T3}) and measurement noise levels (\ttt{S1}-\ttt{S3} \wrt $\xi$). The proposed {HvM} kernel delivers superior tracking accuracy of particle filtering via GP-based reweighting over the PPRD, PvM, PSE kernels, and the parametric modeling method. Exemplary runs of HvM-GP-based particle filtering on these three trajectories are shown in \figref{fig:exa} under the noise level of $\xi=0.01$. Almost identical accuracies are achieved by the PPRD and PvM kernels, which suggests that the von Mises kernel can be interpreted as a reformulation of the common periodic kernel. Moreover, results from these two kernels show better accuracy compared with the parametric method, whereas the PSE kernel fails tracking on most sequences due to disregard of periodic nature of hypertori.
	
	To bring more insights to the functionality of the proposed {HvM} kernel on particle reweighting, we demonstrate in \figref{fig:comp} range measurement distributions predicted by GPs using HvM, PvM, and PSE kernels over time. The proposed HvM-GP predicts more data-adaptive range distributions than PvM-GP, and the {PSE} kernel induces discontinuity when periodicity in inputs takes effect. The parametric method fails at modeling the time-varying noise pattern, which can be interpreted in a straightforward way (thus not plotted).
	
	\section{Conclusions}\label{sec:conclusions}
	We provide a novel study on establishing GPs on the product of directional manifolds. Based on the circular kernel following the form of the von Mises distribution, a novel hypertoroidal von Mises (HvM) kernel has been proposed on the hypertorus $\Sbb^1\times\Sbb^1\times\Sbb^1\subset\R^6$ in a manifold-adaptive manner. It captures information not only within each circular component but also in the correlated region, leading to a more distinct similarity quantification of hypertoroidal data compared to conventional strategies such as kernel multiplication. In consideration of potential usage in runtime-critical scenarios, derivatives of the marginal likelihood \wrt the hyperparameters have been provided to facilitate efficient GP modeling. The proposed HvM-based GP has been evaluated for data-driven recursive localization in ranging-based sensor networks. Simulations have shown that it delivers superior tracking performance over the parametric model and GPs based on kernel multiplications.
	
	For future work, the efficiency of the proposed method can be improved by reducing the computational complexity, e.g., via sparse GPs~\cite{Titsias2009}. We will also present the proof on the positive definiteness of the proposed kernel in a follow-up work. In addition, the proposed HvM-GP is to be exploited in real-world applications for data-driven modeling of complex stochastic systems with multiple angular inputs~\cite{Cai2021}. Also, the HvM kernel can be extended to higher-dimensional spaces, enabling its application in more extensive scenarios.
	
	\section*{Acknowledgment}
	This work was supported in part by the German Research Foundation (DFG) under grant HA 3789/25-1, in part by the Swedish Research Council under grant Scalable Kalman Filters, and in part by ZENITH of Linköping University under grant CASIIS. The work was partially conducted at Karlsruhe Institute of Technology (KIT), Germany.
	\bibliographystyle{IEEEtran}
	\bibliography{Ref.bib,bibOwn.bib}
	
\end{document}